\documentclass[11pt]{article}

\usepackage[preprint]{acl}  % arXiv version; [review] gives the anonymous ARR version
\usepackage{times}
\usepackage{latexsym}
\usepackage[T1]{fontenc}
\usepackage[utf8]{inputenc}
\usepackage{microtype}
\usepackage{inconsolata}
\usepackage{graphicx}
\usepackage{subcaption}
\usepackage{booktabs}
\usepackage{placeins}
\usepackage{amsmath}
\usepackage{amssymb}
\usepackage{algorithm}
\usepackage{algorithmic}
\usepackage{xspace}
\usepackage{mathtools}
\usepackage{amsthm}
\usepackage{pifont}
\usepackage{xcolor}
\usepackage{colortbl}
\usepackage{array}
\usepackage{multirow}
\usepackage{enumitem}
\usepackage[capitalize,noabbrev]{cleveref}

\theoremstyle{plain}

\theoremstyle{definition}

\theoremstyle{remark}

\definecolor{light-purple}{RGB}{151,156,171}
\definecolor{blue-color}{RGB}{40,166,189}
\definecolor{pink-color}{RGB}{237,46,104}
\definecolor{dark-grey-color}{RGB}{79,91,102}
\definecolor{darkbyzantium}{rgb}{0.36, 0.22, 0.33}
\definecolor{bluebell}{rgb}{0.64, 0.64, 0.82}
\definecolor{airforceblue}{rgb}{0.36, 0.54, 0.66}
\newcommand{\toolname}{ContextWeaver\xspace}

\definecolor{posgreen}{RGB}{0,122,61}
\definecolor{negred}{RGB}{190,30,45}
\definecolor{oursbg}{RGB}{232,241,252}
\definecolor{groupbg}{RGB}{243,244,246}

\usepackage[skins,breakable]{tcolorbox}
\usepackage{listings}
\tcbuselibrary{breakable}
\lstdefinestyle{markdown}{
  basicstyle=\ttfamily\small,
  columns=flexible,
  breaklines=true,
  breakatwhitespace=true,
  breakautoindent=true,
  commentstyle=\color{gray},
  keywordstyle=\color{blue}\bfseries,
  stringstyle=\color{purple},
  identifierstyle=\color{black},
  showstringspaces=false
}
\newtcolorbox[auto counter,number within=section]{prompt}[1][]{
    breakable,
    colbacktitle=airforceblue,
    colframe=airforceblue,
    fontupper=\normalsize\raggedright,
    title after break={\thetcbcounter{}~(continued)},
    boxsep=5pt,
    left=0pt,
    right=0pt,
    top=0pt,
    bottom=0pt,
    boxrule=1pt,
    enhanced,
    skin first=enhanced,
    skin middle=enhanced,
    skin last=enhanced,
    #1,
}
\newcommand{\promptsubsection}[1]{
\setlength{\parskip}{6pt} \noindent\textbf{{#1}:}
}
\usepackage[disable,textsize=tiny]{todonotes}

\title{\toolname: Selective and Dependency-Structured \\ Memory Construction for LLM Agents}

\author{Yating Wu$^{1}$\thanks{Corresponding author: \texttt{yating.wu@utexas.edu}}, Yuhao Zhang$^{1}$, Sayan Ghosh$^{1}$, Sourya Basu$^{2}$\thanks{Work done while at Amazon.}, \\
  \textbf{Anoop Deoras$^{1}$, Jun Huan$^{1}$, Gaurav Gupta$^{1}$} \\
  $^{1}$AWS AI Labs \quad $^{2}$Capital One}

\begin{document}
\maketitle

\begin{abstract}
Large language model (LLM) agents often struggle in long-context interactions. As the history grows, sliding windows and prompt compression may drop earlier information that later steps rely on. Retrieval-based memory systems surface relevant content but still overlook the causal and logical dependencies between steps. We introduce \toolname{}, a selective and dependency-structured memory framework that organizes an agent's interaction trace into a graph of reasoning steps and selects the relevant context for future actions. \toolname{} supports (1) dependency-based construction and traversal that link each step to the earlier steps it relies on; (2) compact dependency summaries of the reasoning paths; and (3) a lightweight validation layer that incorporates execution feedback. 
On five software-engineering and terminal benchmarks with a fixed budget per task, \toolname{} improves the resolve rate over a sliding-window baseline on all five, with a significant gain when pooled across benchmarks, while using 13--18\% fewer agent steps. It also performs favorably against alternative context-management approaches, with the largest gains on long-horizon repository tasks. On SWE-Bench Pro, \toolname{} raises the resolve rate from 66.5\% to 72.6\%, with significant improvements across all three model backbones. The improvement also holds when the sliding window's budget is raised to cover the summary cost.

\end{abstract}

\section{Introduction}
Current Large Language Models (LLMs) have a maximum context length, so the model or the underlying infrastructure cannot generate a response when the input exceeds the limit. Even within the limit, a long context can degrade performance \citep{liu2024lost, lilong}. For LLM agents that use tools to iteratively plan and interact with external systems, the context grows especially quickly~\citep{maharana2024evaluating}. It is therefore necessary to selectively compact or reorganize an agent's interaction history.

Approaches such as sliding windows, prompt compression, or retrieval control the context length. Although these methods reduce the number of tokens, they typically select content based on recency, salience, or semantic similarity.
Such signals do not capture the dependency structure that links one reasoning step to the next. In agentic settings, each action can depend on (i) earlier decisions, (ii) tool outputs, and (iii) intermediate hypotheses generated during the trajectory. When these dependencies are lost, the agent can break ongoing plans, repeat earlier exploration, or produce steps that no longer match the earlier context.

In this work, we present \toolname, a selective and dependency-structured memory framework for LLM agents. \toolname consists of three main components:
(1) a dependency-based construction module that links each reasoning step to the earlier steps it relies on;
(2) compact dependency summaries that provide concise, reusable representations of the reasoning paths supporting each step; and
(3) a lightweight validation layer that incorporates execution feedback.
These components let \toolname preserve the context most relevant to the agent's next action while remaining within token budgets.

\begin{figure*}[t]
    \centering
    \includegraphics[width=0.85\textwidth]{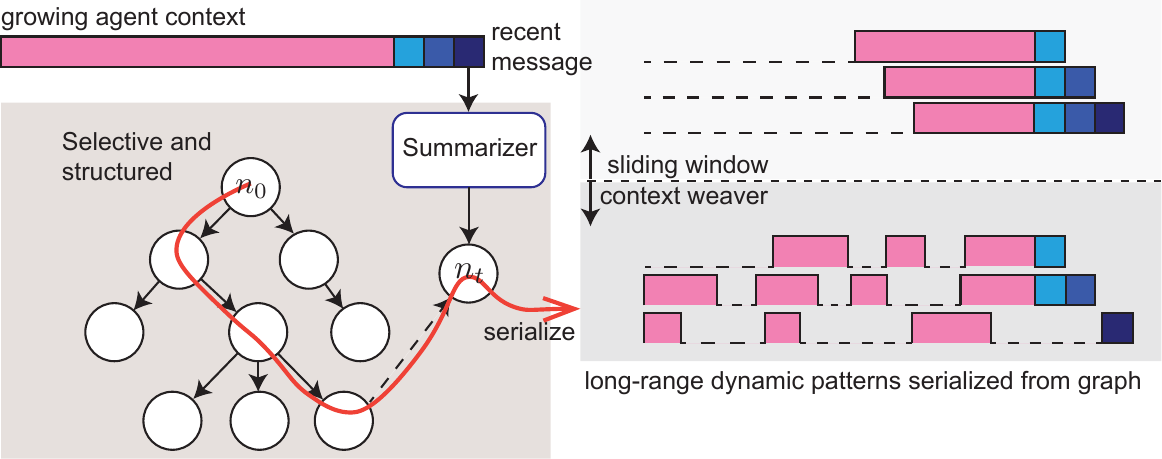}
    \caption{Pipeline of \textsc{ContextWeaver}. AI-agent context grows with environment interactions and tool-calling. The latest message is summarized and converted to graph node $n_t$. The node is linked to the earlier steps it depends on in an iteratively built dependency graph with root ($n_0$) at the initial conversation; a node can have several parents, and the drawing shows one for clarity. \textbf{(Right bottom)}: Following the dependencies of the current step, \toolname{} keeps a few earlier observations in full alongside the most recent steps and replaces the other observations with their node summaries. \textbf{(Right top)}: A sliding window keeps only the most recent observations and drops the older ones. A step-by-step walkthrough of this pipeline on a real SWE-bench Verified instance (\texttt{pytest-dev\_\_pytest-5262}) is provided in Appendix~\ref{prompt:algorithm_walkthrough}.}
    \label{fig:pipeline}
\end{figure*}

Figure~\ref{fig:pipeline} provides an overview of \toolname{}. In contrast to a sliding-window buffer, which only keeps the most recent turns and discards earlier (but essential) information, \toolname{} converts each new reasoning step into a structured node that stores the state of the step (such as action and observation summaries) and identifies the earlier steps that support it. These nodes are linked into a dependency graph that reflects the actual flow of reasoning rather than recency. A summarization module produces compact summaries of the dependency paths leading to the current step, and a validation layer filters out unreliable nodes using execution outcomes. The resulting context keeps validated nodes and their supporting dependencies, which preserves the key evidence for the next action under a fixed budget.

We evaluate \toolname{} across five software-engineering and terminal benchmarks using three model backbones. It consistently improves resolve rates over a sliding-window baseline while requiring fewer agent steps, with the largest gains on long repository tasks that modify multiple files. It also resolves more tasks than keeping the full history, LLM summarization~\citep{wang2025openhands}, and a \emph{hybrid} that adds periodic summarization to a sliding window~\citep{lindenbauer2025complexity}, and more than retrieval on most repository benchmarks. With Kimi-K2.5, the improvements hold even when the sliding window gets the same total budget (Section~\ref{sec:backbone}).

\section{Related Work}
\label{sec:relatedwork}
% \subsection{Prompt Compression}
\paragraph{Prompt Compression}

Existing work on prompt compression aims to reduce token numbers by distilling salient information from original context. Techniques like gist tokens \citep{mu2023learning} offer learned compression context but suffer from limited interpretability and introduce additional training cost. Selective Context \citep{li2023compressing} drops the lexical units with the lowest self-information under a small language model, which need not reflect whether a piece of context will be needed by later reasoning steps. Token-level methods such as LLMLingua \citep{jiang2023llmlingua} and embedding-level compression methods \citep{chevalier2023adapting} also fall into this category. 

% \subsection{Retrieval and Hierarchical Memory}
\paragraph{Retrieval and Hierarchical Memory}
Related systems such as MemGPT \citep{packer2023memgpt}, Generative Agents \citep{park2023generative}, and Reflexion \citep{shinn2023reflexion} store past interactions in hierarchical or database-like memories. While these approaches scale to longer contexts, retrieval is largely driven by semantic similarity, recency, or reflective signals. Furthermore, they still operate over linearized token streams once retrieved, without explicitly modeling logical or causal relations between interaction steps. More recent agent memories such as Mem0~\citep{chhikara2025mem0} extract and consolidate salient facts across sessions, in one variant as a graph of related entities; they target long-term knowledge about users and tasks rather than the observations an agent needs within a single ongoing trajectory.
% \subsection{Graph-Based Reasoning vs. Memory}
\paragraph{Graph-Based Reasoning vs. Memory}
Recent work on graph-based memory organization primarily focuses on graph construction and retrieval \citep{xu2025mem}, with less attention to how these structures can be simplified or optimized over time. Building on Chain-of-Thought prompting \citep{wei2022chain}, graph-based reasoning frameworks such as Tree of Thoughts (ToT) \citep{yao2023tree} and Graph of Thoughts (GoT) \citep{besta2024graph} study how structured representations can support inference-time search by branching over alternative intermediate states. Unlike these approaches, which organize alternative reasoning states during inference, \toolname{} constructs a graph over steps the agent has already executed. The edges record dependencies between earlier observations and subsequent actions, allowing the agent to recover relevant context as the task progresses.

% \subsection{Static Analysis and Repository Context}
\paragraph{Static Analysis and Repository Context}
\label{sec:code_context}
While general-purpose memory systems focus on text, specialized software engineering agents often rely on static analysis to manage context. Agent platforms such as OpenHands \citep{wang2025openhands} and non-agentic pipelines such as Agentless \citep{xia2025agentless} retrieve over entire code repositories \citep{liu2023repobench} and use structural representations such as Abstract Syntax Trees (ASTs) to capture code dependencies. These methods are inherently \textit{static}. They describe code structure but do not model the \textit{dynamic} progression of the agent’s debugging process. In tasks such as SWE-Bench \citep{jimenez2023swe}, critical context often appears in runtime error logs, test failure messages, or temporary variable values, rather than in static source files. ContextWeaver complements static analysis by modeling the logical dependencies in the \textit{interaction history}, preserving the reasoning path that leads to a code modification.

\paragraph{LLM Summarization}
LLM-based summarization periodically condenses older agent interactions to control context length. OpenHands condenses its event histories this way~\citep{wang2025openhands, wang2025openhandssdk}, and ACON~\citep{kang2025acon} optimizes the compression guidelines themselves, while SWE-agent masks older observations by default~\citep{yang2024swe}. \citet{lindenbauer2025complexity} find that such observation masking can match LLM summarization at substantially lower cost and propose a hybrid of the two, which we include as a baseline. Both summarization and masking reduce context length without tracking which earlier observations subsequent steps depend on.
\section{\toolname: A Dependency-Structured Memory Framework}
\label{sec:contextweaver}
We present \textbf{\toolname}, a framework that organizes agent interaction histories by maintaining the causal and logical structure in a dependency graph. Figure~\ref{fig:pipeline} summarizes the overall pipeline. 
The key component of \toolname{} is dependency-aware context construction (Section~\ref{subsec:algorithm}), which maintains a dependency graph based on logical and causal relationships and weaves context for the next decision of the LLM agent.
Section~\ref{sec:depsum} introduces the Dependency Summarizer, which produces compact summaries of dependency paths to guide context selection under token constraints.
For each step, a Validation Layer (Section~\ref{sec:validation}) records its execution outcome and annotates it with a reliability signal to inform context weaving in later steps.

\label{subsec:problem}
\paragraph{Problem Setup}
Tool use is central to many agent settings, where tasks depend on external operations such as web search, code execution, or data retrieval \citep{schick2023toolformer, gao2023pal}. Each tool invocation introduces new observations and implicit causal links between steps, creating a reasoning trajectory that is not captured by standard conversational histories~\citep{yao2023react}. 
We focus on the histories of tool-using LLM agents, rather than simple chat agents whose histories consist primarily of user--assistant message pairs.

We represent the history of a tool-using agent as
\[
\begin{aligned}
\mathbf{H} = \{&\{\text{system}, Q\}, \{T_1, A_1, O_1\},\\
&\dots, \{T_n, A_n, O_n\}\},
\end{aligned}
\]
where $Q$ denotes the user query, "system" denotes the system prompt, and each triple $(T_i, A_i, O_i)$ defines a single reasoning step: the \textbf{Thought} $T_i$ captures the agent's internal reasoning, the \textbf{Action} $A_i$ specifies the tool invocation, and the \textbf{Observation} $O_i$ records the resulting tool output.
\toolname{} extracts these components into structured nodes and constructs a dependency graph over reasoning steps. It then generates the final context passed to the LLM agent for the next decision.
% \sayan{This para should conclude by saying ContextWeaver generates the final context for the next call of LLM agent. Right now, this para only ends on the graph construction part.}

\subsection{Dependency-Aware Context Construction}
\label{subsec:algorithm}
Algorithm~\ref{alg:contextweaver} (Appendix~\ref{appendix:walkthrough}) presents \toolname{} as a framework that takes an agent’s interaction history and a user query as input, and produces a compact, dependency-aware context as output. Given a history $\mathbf{H}$ and query $Q$, the method constructs a structured representation $N_k$ of the most recent step $H_k$, links $N_k$ to its dependent earlier steps $S_k$, and extracts only the validated dependencies needed for the agent’s next action. The algorithm proceeds as follows: (1) \textbf{Node Extraction}: converting the latest history entry into a structured node $N_k$, (2) \textbf{Parent Selection}: identifying its parent dependencies $S_k$ and adding edges $N_i \rightarrow N_k (\forall N_i \in S_k)$ to the dependency graph, (3) \textbf{Ancestry Dependency Construction}: retrieving the dependency subgraph from the identified ancestry dependency $A$, and (4) \textbf{Context Weaving}: constructing the final set of context $C$ according to $A$. The hyperparameter $W$ sets the warmup period and bounds the size of the ancestry subgraph.
% This process runs incrementally as the agent interacts with tools, allowing the context structure to evolve online without retraining or reprocessing the full history.
% TODO: if validation filtering is added back, restore step (4) filtering invalid or superseded nodes in $P_k$ to obtain a validated subgraph $P_k^f$, and renumber (5)

\toolname{} operates in a local, incremental manner -- instead of reprocessing the full history, it extends the dependency graph with new nodes and selectively updates node validity through the Validation and Test Layer (Section~\ref{sec:validation}).
% \subsection{Dependency-Aware Context Construction}
% \label{sec:graph}
% To construct a dependency-aware representation, \toolname begins with node extraction, which summarizes the new step and initializes its fields. It then proceeds to parent selection and ancestry dependency construction, which determine how the new node connects to prior reasoning and which parts of history have the causal and logical dependency with the current step.
% Finally, it weaves the context by compressing the rest of the history that does not support this step.

\paragraph{Node Extraction}
\label{subsec:trace_extractor}
Given the most recent tool-use step $H_k = \{T_k, A_k, O_k\}$, \toolname converts it into a \textit{Node} $N_k$. 
\[
\begin{aligned}
N_k = (&\texttt{summary}, \texttt{dependency\_summary},\\
&\texttt{parents}, \texttt{validation}).
\end{aligned}
\]
 We denote the node extraction mapping by $\phi(H_k, Q)$, which summarizes
  the latest step given the query.
Here, \texttt{summary} captures the agent’s thought, action, and resulting observation, and
\texttt{dependency\_summary} (introduced in Section~\ref{sec:depsum}) provides a concise narrative of the reasoning dependency paths leading to the node.
Also, \texttt{parents} encodes its connections within the directed acyclic graph (DAG), while  
\texttt{validation} records node-level outcomes (e.g., passed, failed, unknown or superseded) and includes task-specific metadata such as test outputs (Section~\ref{sec:validation}).  

When a new reasoning step is extracted as a node $N_k$, \toolname
incorporates it into a dependency graph that represents the
logical and causal structure of the agent's history.  The graph
organizes operations not by recency but by information flow, forming a
 DAG that supports branching and merging across
parallel lines of reasoning.

 \paragraph{Parent Selection}
  For each new node $N_k$, \toolname identifies its supporting predecessors
  using an LLM-based \textit{Logical Dependency Analyzer} (Appendix~\ref{prompt:memory_weaver_parent}). Let $
  \mathcal{C}_k$ denote the valid nodes in the dependency
  graph, and the analyzer returns up to $m$ parents:
  \[
  S_k = \underset{N_i \in \mathcal{C}_k}{\mathrm{Top_m}}\ \mathrm{LLM}(N_i \to N_k \mid Q).
  \]
  The LLM-based analyzer scores each node $N_i$ by
  $\mathrm{LLM}(N_i \to N_k \mid Q)$, which is the estimated likelihood that $N_k$ depends on $N_i$. It then returns the top-$m$ parents $S_k$. Because a node may have several
  parents, a step can build on information from several earlier steps rather than a single chain.

% \sayan{Mention the function name "SearchParents" here otherwise its an unreferenced/un=explained variable in equation}
Parent selection is treated as a reasoning task rather than a simple similarity check. 
% The procedure first narrows the list of possible predecessors. 
For each candidate, the analyzer reasons whether the current step depends on earlier information. For example, the dependency may involve using a prior result, returning to an earlier issue, or continuing a line of reasoning. The analyzer then selects the candidates that support the new step. Our experiments replace the analyzer with a cheaper score (see \emph{Low-cost configuration} below). 

\paragraph{Ancestry Dependency Construction}
\label{analysis:chain_summary_build}

After identifying the parents of $N_k$, \toolname collects the prior reasoning steps that causally or logically contribute to this node by constructing the ancestry set $A$, defined as all nodes that directly or indirectly support $N_k$. 
This is done via a breadth-first traversal upward through the dependency graph, starting with ${N_k}$ and expanding along parent edges. 
The traversal continues until either the queue is exhausted or $A$ holds $W$ ancestors besides $N_k$. The selected ancestors determine the final context $C$, which focuses on the most relevant history entries.

\paragraph{Context Weaving}
% \label{sec:weaving}
\toolname{} weaves the final context $C$ according to the ancestry subgraph $A$ of the current node. It retains the full \textbf{T--A--O} triples for nodes within this subgraph, and for all other history entries, preserves the \textbf{Thought} ($T_i$) and \textbf{Action} ($A_i$) while compressing the corresponding \textbf{Observation} ($O_i$) by replacing it with a placeholder indicating omitted content (Appendix~\ref{appendix:observation-compress-sec}).

% \paragraph{Outcome}
% The resulting DAG provides (i) a principled way to retain only the
% \textit{supporting} steps for the current goal via ancestry selection, and
% (ii) a compact, human-interpretable account of how the agent arrived at the current node
% $N_k$. This graph is then used by the validation-aware context
% selector (Section~\ref{sec:weaving}). 
% We leave node merging/folding as future work 

\subsection{Dependency Summarizer}
\label{sec:depsum}
For interpretability, each node maintains a brief \textit{dependency summary} that records the main reasoning dependency from the root to that node. \toolname{} constructs this summary incrementally by providing the dependency summaries of parents and the current node’s summary, rather than re-summarizing the full history at every step. Denote the LLM summarizer as $\mathrm{LLMSUM}$, the summary of $N_k$ as $\texttt{summary}_k$, and its dependency summary as $\texttt{dep\_sum}_k$; then
\[
\begin{aligned}
\texttt{dep\_sum}_k = \mathrm{LLMSUM}\big(&\{\texttt{dep\_sum}_i \mid N_i \in S_k\},\\
&\texttt{summary}_k\big).
\end{aligned}
\]

\subsection{Validation and Test Layer} 
\label{sec:validation}
To incorporate execution feedback, \toolname adds a 
validation layer that records outcomes from each testing or verification
step. This layer provides two complementary signals that operate at
different levels of granularity.

\paragraph{Fine-grained evidence.}
A \textit{TestTracker} parses tool observations such as
\texttt{pytest} or \texttt{unittest} outputs and extracts individual test
results into the field \texttt{test\_results}. These entries capture which
tests passed or failed and populate the global tracker used to generate the
prepended test summary. They also support cross-referencing between
nodes: when a later node fixes an earlier failure, the system links them
through a \texttt{superseded} pointer.

\paragraph{Node-level abstraction.}
Based on the collected test results, the \textit{ValidationDetector}
assigns each node a simple label,
 \texttt{validation\_status}\allowbreak~$\in\{\text{passed},\text{failed},
  \text{unknown},\text{superseded}\}$, to indicate whether the step produced a stable outcome. During graph
construction, nodes marked as \texttt{failed} are skipped when searching
for parent candidates, so new steps never depend on incomplete or broken
states. Labels come only from test-suite runs (\texttt{pytest} or \texttt{unittest} output); other steps, such as a reproduction script that confirms the bug, stay \texttt{unknown}. Nodes labeled as \texttt{passed} or \texttt{unknown} remain
eligible for parent selection. Nodes marked \texttt{superseded} are similarly skipped, as their content has been replaced by a later correction. The status labels allow parent selection to filter nodes without re-parsing their test logs.

\paragraph{Low-cost configuration.}
\label{sec:lowcost}
Our experiments use a lower-cost instantiation of the framework: Claude Haiku 4.5 writes the node and dependency summaries, parents are scored by shared files, shared identifiers and recency rather than by an LLM call, and the current step and the two before it are always kept in full. Observations outside the ancestry set are replaced by their node summary (at most 300 characters) rather than a placeholder. We chose this configuration on a small development subset of SWE-Bench Verified and the 26-task Pro sample and kept it fixed for every benchmark and backbone; excluding those 26 tasks leaves the Pro Python result unchanged (Appendix~\ref{appendix:additional_benchmarks}). We also evaluate the full framework, which uses an LLM-based parent analyzer and traverses dependencies over multiple hops; on the 186 SWE-Bench Pro Python tasks, it performs on par with the low-cost configuration with both Kimi-K2.5 and Claude Sonnet 4.5 (Table~\ref{tab:full_variant}).

\section{Experimental Setup}
\label{sec:setup}
\label{sec:experiments}

\paragraph{Benchmarks.}
We focus on long-horizon tasks, where an agent's history outgrows a sliding window: repository tasks with multi-file or release-level changes, and terminal tasks. We use five benchmarks (details in Appendix~\ref{appendix:additional_benchmarks}).
\begin{itemize}[leftmargin=*,itemsep=0pt,topsep=2pt]
    \item \textbf{SWE-Bench Pro}~\citep{deng2025swebenchpro}: its public release (642 tasks in 11 repositories at the time of our experiments; the commercial and held-out sets are not public). We use all 237 Python tasks, including a 26-task \emph{Pro sample} used during development; a \emph{large-change} subset of 30 Python tasks whose gold patch touches at least 4 files or 150 lines; the benchmark's 51 hardest tasks; and the 344 Go/JS/TS tasks outside those 51 and a 30-task pilot.
    \item \textbf{SWE-Bench Multilingual}~\citep{yang2025swesmith}: all 300 tasks in 9 languages.
    \item \textbf{SWE-EVO}~\citep{le2025sweevo}: all 48 release-level tasks.
    \item \textbf{SWE-rebench}~\citep{badertdinov2025swerebench}: the 216 tasks of its 860-task leaderboard set whose gold patch touches at least 3 files or 100 lines.
    \item \textbf{Terminal-Bench 2.0}~\citep{tbench2025}: all 89 tasks.
\end{itemize}
A task counts only if it was graded in every run of both methods ($n$ in the tables). Tasks drop out when their environment fails to build or, on SWE-rebench and SWE-EVO, when the gold patch fails the benchmark's own tests.

\paragraph{Agent, models and budget.}
All agents use \textsc{SWE-agent}~\citep{yang2024swe} with the same system prompt, tools and instance template (Appendix~\ref{prompt:agent_setup}). The agent model is Claude Sonnet 4.5, with a cap of \$3 per task on the agent model and at most 150 steps. As further backbones, we use the open-weight Kimi-K2.5 and GLM-5 with the same token budget: their prices per input token are one fifth and one third of Sonnet's, so their caps are \$0.60 and \$1.00. \toolname{}'s summaries use Claude Haiku 4.5 with every backbone. Because these summaries add cost, we also run the sliding window with Kimi-K2.5 under the same total budget as \toolname{}, raising its cap from \$0.60 to \$1.08: the agent cap plus \toolname{}'s average summary cost per task (\$0.48). Both methods use the same test-based stopping check: when a patch can be submitted, the check runs the repository tests that reference the changed functions before and after the patch, and stops the run if the agent's reproduction passes and no such test newly fails.

\paragraph{Baselines.}
The main baseline is a sliding window that keeps the $n{=}6$ most recent observations, as many as \toolname{} keeps in full. We also compare with four other ways of managing the history: keeping all of it; LLM summarization in the style of the OpenHands condenser~\citep{wang2025openhands}, which condenses older history with the agent model once more than 10 observations have accumulated (Appendix~\ref{appendix:llm_summarization}); the \emph{hybrid} of \citet{lindenbauer2025complexity}, which keeps a sliding window of the last 10 observations and lets the agent model summarize the oldest 43 turns whenever 53 have accumulated; and \emph{retrieval}, which keeps six observations in full like the sliding window but picks three of them among the older ones by BM25 relevance to the current step, the sparse retriever that SWE-bench uses to select code context~\citep{jimenez2023swe}. We run these four on the Pro sample and the large-change subset; retrieval and full history also run on longer task sets (Appendix~\ref{appendix:additional_benchmarks}).

\paragraph{Metrics and statistics.}
We report the resolve rate under each benchmark's own tests. Most configurations are run three times. We compare the two methods task by task with an exact McNemar test. Across benchmarks, we also use a sign test with each task set as one observation, which does not assume that task runs are independent.

\begin{table}[t]
\centering
\small
\setlength{\tabcolsep}{3.1pt}
\resizebox{\columnwidth}{!}{%
\begin{tabular}{@{}l r r@{\,}l r@{\,}l r@{}}
\toprule
\textbf{Benchmark} & \textbf{$n$} & \multicolumn{2}{c}{\textbf{SW}} & \multicolumn{2}{c}{\textbf{CW (ours)}} & \textbf{$\Delta$} \\
\midrule
\multicolumn{7}{@{}l}{\textit{SWE-Bench Pro}} \\
\quad Python & 186 & 66.5 & {\scriptsize$\pm$2.2} & \textbf{72.6} & {\scriptsize$\pm$1.4} & \textbf{+6.1}$^{***}$ \\
\quad Large-change & 22 & 60.6 & {\scriptsize$\pm$5.2} & \textbf{71.2} & {\scriptsize$\pm$6.9} & +10.6 \\
\quad Go/JS/TS & 175 & 37.5 & {\scriptsize$\pm$2.8} & \textbf{41.3} & {\scriptsize$\pm$1.2} & +3.8 \\
\midrule
\multicolumn{7}{@{}l}{\textit{Repository-level tasks}} \\
\quad Multilingual & 262--289$^{\ddagger}$ & 43.2 & {\scriptsize$\pm$1.6} & \textbf{46.7} & {\scriptsize$\pm$1.6} & \textbf{+3.5}$^{*}$ \\
\quad SWE-EVO & 42 & 23.8 & {\scriptsize$\pm$2.4} & \textbf{31.7} & {\scriptsize$\pm$2.7} & +7.9 \\
\quad SWE-rebench & 212 & 27.2 & {\scriptsize$\pm$1.0} & \textbf{30.3} & {\scriptsize$\pm$0.3} & \textbf{+3.1}$^{*}$ \\
\midrule
\multicolumn{7}{@{}l}{\textit{Terminal tasks}} \\
\quad Terminal-Bench 2.0 & 61 & 43.2 & {\scriptsize$\pm$4.1} & \textbf{44.8} & {\scriptsize$\pm$0.9} & +1.6 \\
\midrule
\textbf{Macro average} & & 40.8 & & \textbf{45.2} & & \textbf{+4.4} \\
\bottomrule
\end{tabular}%
}
\caption{Resolve rate (\%) with Claude Sonnet 4.5 and a \$3 cap per task: mean $\pm$ std over three runs. $n$: tasks graded in every run of both methods ($^{\ddagger}$Multilingual: 262--289 per run). Stars give an exact McNemar test on paired task outcomes, pooled over runs (which treats repeated runs as independent pairs, so the $p$-values are approximate): $^{*}p{<}0.05$, $^{***}p{<}0.001$. Macro average: one row per benchmark (Pro Python, Multilingual, SWE-EVO, SWE-rebench, Terminal-Bench). SW: sliding window ($n{=}6$); CW: \toolname{}. Per-run results are in Appendix~\ref{appendix:additional_benchmarks}.}
\label{tab:main}
\end{table}

\begin{figure}[t]
\centering
\includegraphics[width=\columnwidth]{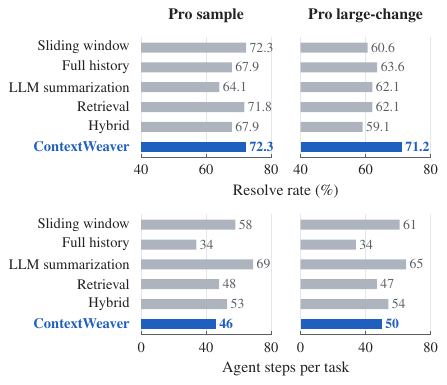}
\caption{Resolve rate (top) and agent steps per task (bottom) of \toolname{} and the other context-management methods on two Pro task sets (26 and 22 tasks; Claude Sonnet 4.5, \$3 cap; mean over three to five runs). Every method has the same budget. Standard deviations and costs per task are in Table~\ref{tab:baselines_full}.}
\label{fig:baselines}
\end{figure}

\begin{table}[t]
\centering
\small
\setlength{\tabcolsep}{3.5pt}
\resizebox{\columnwidth}{!}{%
\begin{tabular}{@{}l c r@{\,}l r r@{}}
\toprule
\textbf{Method} & \textbf{Runs} & \multicolumn{2}{c}{\textbf{Resolve (\%)}} & \textbf{Steps} & \textbf{\$/task} \\
\midrule
\multicolumn{6}{@{}l}{\textit{Kimi-K2.5: Pro Python ($n$=186), cap \$0.60}} \\
Sliding window & 3 & 54.1 & {\scriptsize$\pm$1.9} & 70 & 0.45 \\
Sliding window (cap \$1.08) & 3 & 53.4 & {\scriptsize$\pm$1.9} & 88 & 0.65 \\
\toolname{} & 3 & \textbf{62.7}$^{***\dagger}$ & {\scriptsize$\pm$3.2} & \textbf{55} & 0.91 \\
\midrule
\multicolumn{6}{@{}l}{\textit{Kimi-K2.5: Verified ($n$=98)}} \\
Sliding window & 3 & 65.6 & {\scriptsize$\pm$4.1} & 51 & 0.31 \\
\toolname{} & 3 & \textbf{68.7} & {\scriptsize$\pm$2.6} & \textbf{45} & 0.68 \\
\midrule
\multicolumn{6}{@{}l}{\textit{GLM-5: Pro Python ($n$=186), cap \$1.00}} \\
Sliding window & 3 & 61.6 & {\scriptsize$\pm$3.8} & 75 & 0.77 \\
\toolname{} & 3 & \textbf{69.5}$^{***}$ & {\scriptsize$\pm$3.7} & \textbf{58} & 1.27 \\
\bottomrule
\end{tabular}%
}
\caption{Other backbones, and a sliding window with the same total budget as \toolname{} (cap \$1.08, the \$0.60 agent cap plus \toolname{}'s average summary cost). Resolve rate is the mean $\pm$ std over runs; for resolve rate and steps, best in \textbf{bold}. $^{*}$, $^{**}$, $^{***}$: $p{<}0.05$, $0.01$, $0.001$ against the sliding window; $^{\dagger}$: $p{<}0.05$ against the sliding window with the \$1.08 cap (exact McNemar, pooled over runs). \$/task is the total cost, including \toolname{}'s Haiku summaries.}
\label{tab:backbone_baselines}
\end{table}

\begin{figure*}[t]
\centering
\includegraphics[width=\textwidth]{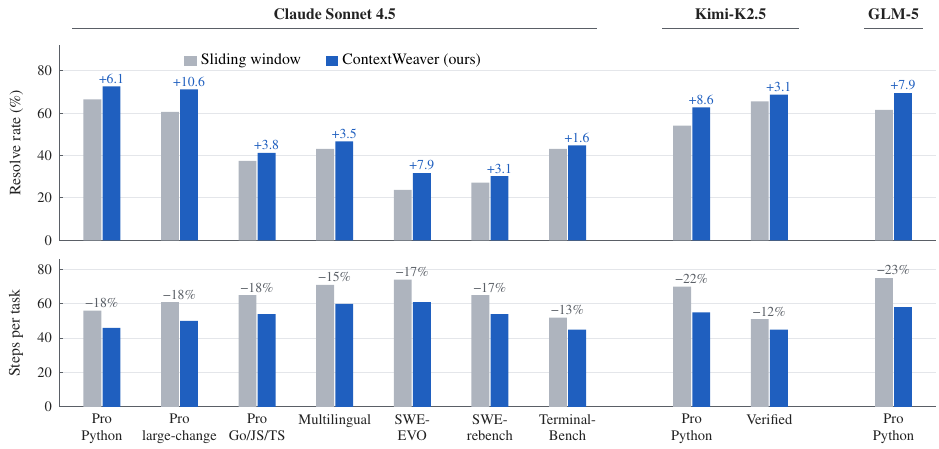}
\caption{Resolve rate (top) and agent steps per task (bottom) of the sliding window and \toolname{} on each benchmark and backbone; the numbers above the bars give \toolname{}'s gain in resolve rate and its change in steps. Steps are averaged over all task runs. Many runs of both methods stop when the budget is used up, and a \toolname{} step costs more because its prompt is longer, so part of the step reduction reflects the budget; on SWE-Bench Pro task runs that both methods solve before the budget runs out, \toolname{} still needs 9--18\% fewer steps (Table~\ref{tab:steps_budget}).}
\label{fig:efficiency}
\end{figure*}

\section{Results}
\label{sec:results}

\subsection{Main Results}
Table~\ref{tab:main} shows that \toolname{} outperforms the sliding-window baseline on all five benchmarks under the same budget for the agent model. The largest gains are on SWE-Bench Pro, where \toolname{} raises the resolve rate from 66.5\% to 72.6\% on Python tasks and gains most on the large-change subset; over all Pro tasks, including Go, JavaScript and TypeScript, the gain is significant. These gains extend across the remaining benchmarks and to most languages and repositories (Appendix~\ref{appendix:additional_benchmarks}).

\subsection{Comparison with Other Context Management}
\label{sec:baselines}
We compare \toolname{} with full-history retention, LLM summarization, retrieval, and the hybrid on two Pro task sets (Figure~\ref{fig:baselines}). \toolname{} achieves the highest resolve rate on both, tied with the sliding window on the Pro sample. Full-history retention helps on longer tasks, but the growing prompts quickly exhaust the budget. It performs below \toolname{} on all three task sets evaluated, including SWE-EVO (Appendix~\ref{appendix:additional_benchmarks}). LLM summarization requires the most agent steps and costs the most per task. Retrieval performs well on Terminal-Bench and SWE-EVO, while \toolname{} outperforms it on the other repository tasks (Table~\ref{tab:retrieval}). These results highlight the importance of selecting the observations needed for subsequent actions, beyond simply retaining or compressing the history.

\subsection{Agent Steps}
\label{sec:steps}
As Figure~\ref{fig:efficiency} shows, \toolname{} needs 12--23\% fewer steps than the sliding window on every benchmark and backbone. Part of this gap comes from runs that end on the budget, where its longer prompts make each step costlier, but the saving does not depend on the budget: on task runs that both methods solve before the budget runs out, it still needs 9--18\% fewer steps on SWE-Bench Pro with all three backbones (Table~\ref{tab:steps_budget}). Its steps are fewer but longer, so the agent uses about as many tokens per task as with the sliding window.

\subsection{Other Backbones and Equal Total Budget}
\label{sec:backbone}
\toolname{} also improves resolve rates with both open-weight backbones (Table~\ref{tab:backbone_baselines}). Since the summaries add cost, we also compare against a sliding window with the same total budget, using Kimi-K2.5. \toolname{} retains a significant advantage in this comparison, suggesting that the improvement is not simply due to additional model spending.

\begin{table}[t]
\centering
\small
\setlength{\tabcolsep}{3.4pt}
\begin{tabular}{@{}l c c c c@{}}
\toprule
 & & & \multicolumn{2}{c}{\textbf{\toolname{} parents}} \\
\cmidrule(l){4-5}
\textbf{Backbone} & \textbf{Runs} & \textbf{SW} & \textbf{Rule} & \textbf{LLM} \\
\midrule
Kimi-K2.5 & 3 & 54.1 & 62.7 & 63.3 \\
Claude Sonnet 4.5 & 3 & 66.5 & 72.6 & 72.9 \\
\bottomrule
\end{tabular}
\caption{Resolve rate (\%) on the 186 Pro Python tasks with rule-scored parents (used in all other experiments) and LLM-chosen parents (the full framework, Section~\ref{sec:lowcost}). With Sonnet, the two tie ($p{=}0.91$). SW: sliding window.}
\label{tab:full_variant}
\end{table}

\subsection{When Does \toolname{} Help?}
\label{sec:when}
\paragraph{Long tasks with limited budgets.}
The largest gains in Table~\ref{tab:main} are on the Pro large-change subset and on SWE-EVO, the longest tasks. When the budget is tight relative to the task, the steps \toolname{} saves turn into additional attempts.

\begin{figure}[t]
\centering
\includegraphics[width=\columnwidth]{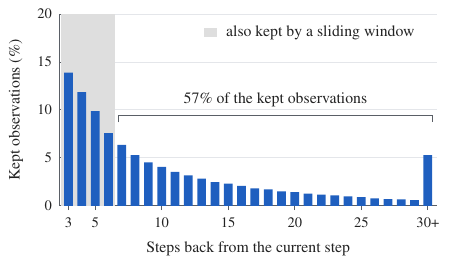}
\caption{Age of the observations that \toolname{}'s graph keeps in full, in steps back from the current step (186 Pro Python tasks, run 1, Claude Sonnet 4.5 and Kimi-K2.5 pooled; 30+ pools older steps). The current step and the two before it are always kept and are not counted. The grey band marks the steps that a six-observation sliding window keeps as well.}
\label{fig:kept_ages}
\end{figure}

\paragraph{What the graph keeps.}
With both backbones, more than half of the observations that \toolname{} keeps in full beyond the two most recent steps are more than six steps old, so a six-observation sliding window would already have dropped them (Figure~\ref{fig:kept_ages}).

\paragraph{Which steps to keep.}
Besides the current step and the two before it, \toolname{} keeps three earlier observations in full, chosen through the dependency graph. To test whether the dependency structure itself matters, we keep everything else fixed and change only how these three are chosen (Table~\ref{tab:selection}). On SWE-EVO, choosing the most recent or random earlier steps lowers the resolve rate from 31.7\% to about 20\%, below the sliding window, even though the node summaries are unchanged. Retrieval, which also picks earlier observations by their relevance to the current step, does about as well as \toolname{} there. Choosing steps by how they relate to the current step, rather than by recency, is therefore essential on these long tasks. On the Multilingual sample, \toolname{} is the best variant, while recency selection falls slightly below the sliding window.

\begin{table}[t]
\centering
\small
\setlength{\tabcolsep}{3pt}
\begin{tabular}{@{}l c c@{}}
\toprule
\textbf{Method} & \textbf{SWE-EVO} & \textbf{Multilingual (50)} \\
\midrule
Sliding window & 23.8 & 40.4 \\
Retrieval (BM25) & \textbf{32.5} & 37.6 \\
Most recent steps & 19.0 & 39.7 \\
Random earlier steps & 22.2 & \underline{43.3} \\
\toolname{} & \underline{31.7} & \textbf{46.8} \\
\bottomrule
\end{tabular}
\caption{Resolve rate (\%, mean of three runs) on the 42 and 47 tasks graded in every run of every method when \toolname{} chooses its earlier full observations by recency or at random instead of by dependency; best in \textbf{bold}, second \underline{underlined}. On SWE-EVO, \toolname{} is significantly better than both ($p{=}0.002$ and $p{=}0.012$).}
\label{tab:selection}
\end{table}

\paragraph{Less re-reading.}
\toolname{} reduces the number of repeated read commands by about 50\% on the 186 Pro Python tasks with both backbones. The case studies in Appendix~\ref{sec:case_study} illustrate this pattern: once an observation leaves the window, the sliding-window agent reads the file again, while \toolname{} keeps the observation and moves on to the edit.

\section{Conclusion}
We presented \toolname{}, which uses dependencies between agent steps to decide which earlier observations to retain. Across five software-engineering and terminal benchmarks and three backbones, \toolname{} improves resolve rates over a sliding-window baseline while using fewer agent steps. On SWE-Bench Pro, it also outperforms full-history retention, LLM summarization, and the hybrid. The largest gains occur on long-horizon repository tasks, where agents need information from earlier in their interaction history. Most of the earlier observations selected by \toolname{} would have been discarded by the sliding window. Selecting these observations by recency or at random substantially reduces performance on the longest tasks, highlighting the importance of choosing them by their relation to the current step. Future work includes improving parent selection with test signals, reducing summarization latency, and extending the framework to multi-agent settings.

\section*{Limitations}
Our evaluation focuses on software-engineering and terminal tasks, where tests provide explicit feedback. Extending \toolname{} to open-ended domains without such feedback remains future work. Its benefits are most pronounced on long tasks under limited budgets; on shorter tasks or with larger budgets, improvements in resolve rate are less consistent. Summarization and longer prompts add computational cost, particularly with cheaper agent backbones. Waiting for a summary at each step also increases wall-clock time. Finally, context selection depends on the quality of parent links and node summaries: missing a dependency can cause an observation needed later to be compressed.

\section*{Ethical Considerations}
All experiments use publicly available benchmarks and open-source repositories, with agents running in isolated containers. No human subjects are involved. Generative AI tools were used to assist with software debugging, consistency checks between the code and manuscript, and language editing. The authors reviewed and verified all AI-assisted text, code, analyses, and claims and take responsibility for the paper's contents.

\bibliography{custom}

@article{mu2023learning,
  title={Learning to compress prompts with gist tokens},
  author={Mu, Jesse and Li, Xiang and Goodman, Noah},
  journal={Advances in Neural Information Processing Systems},
  volume={36},
  pages={19327--19352},
  year={2023}
}

@inproceedings{li2023compressing,
  title={Compressing context to enhance inference efficiency of large language models},
  author={Li, Yucheng and Dong, Bo and Guerin, Frank and Lin, Chenghua},
  booktitle={Proceedings of the 2023 conference on empirical methods in natural language processing},
  pages={6342--6353},
  year={2023}
}

@article{xu2025mem,
  title={A-mem: Agentic memory for llm agents},
  author={Xu, Wujiang and Liang, Zujie and Mei, Kai and Gao, Hang and Tan, Juntao and Zhang, Yongfeng},
  journal={Advances in Neural Information Processing Systems},
  volume={38},
  pages={17577--17604},
  year={2026}
}

@article{liu2024lost,
  title={Lost in the middle: How language models use long contexts},
  author={Liu, Nelson F and Lin, Kevin and Hewitt, John and Paranjape, Ashwin and Bevilacqua, Michele and Petroni, Fabio and Liang, Percy},
  journal={Transactions of the association for computational linguistics},
  volume={12},
  pages={157--173},
  year={2024}
}

@article{lilong,
  title={Long-context llms struggle with long in-context learning},
  author={Li, Tianle and Zhang, Ge and Do, Quy Duc and Yue, Xiang and Chen, Wenhu},
  journal={arXiv preprint arXiv:2404.02060},
  year={2024}
}

@inproceedings{maharana2024evaluating,
  title={Evaluating very long-term conversational memory of llm agents},
  author={Maharana, Adyasha and Lee, Dong-Ho and Tulyakov, Sergey and Bansal, Mohit and Barbieri, Francesco and Fang, Yuwei},
  booktitle={Proceedings of the 62nd Annual Meeting of the Association for Computational Linguistics (Volume 1: Long Papers)},
  pages={13851--13870},
  year={2024}
}

@article{schick2023toolformer,
  title={Toolformer: Language models can teach themselves to use tools},
  author={Schick, Timo and Dwivedi-Yu, Jane and Dess{\`\i}, Roberto and Raileanu, Roberta and Lomeli, Maria and Hambro, Eric and Zettlemoyer, Luke and Cancedda, Nicola and Scialom, Thomas},
  journal={Advances in neural information processing systems},
  volume={36},
  pages={68539--68551},
  year={2023}
}

@inproceedings{gao2023pal,
  title={{PAL}: Program-aided language models},
  author={Gao, Luyu and Madaan, Aman and Zhou, Shuyan and Alon, Uri and Liu, Pengfei and Yang, Yiming and Callan, Jamie and Neubig, Graham},
  booktitle={International conference on machine learning},
  pages={10764--10799},
  year={2023},
  organization={PMLR}
}

@inproceedings{yao2023react,
  title={{ReAct}: Synergizing reasoning and acting in language models},
  author={Yao, Shunyu and Zhao, Jeffrey and Yu, Dian and Du, Nan and Shafran, Izhak and Narasimhan, Karthik R and Cao, Yuan},
  booktitle={The eleventh international conference on learning representations},
  year={2022}
}

@inproceedings{jimenez2023swe,
  title={Swe-bench: Can language models resolve real-world github issues?},
  author={Jimenez, Carlos E and Yang, John and Wettig, Alexander and Yao, Shunyu and Pei, Kexin and Press, Ofir and Narasimhan, Karthik},
  booktitle={International Conference on Learning Representations},
  volume={2024},
  pages={54107--54157},
  year={2024}
}

@article{yang2024swe,
  title={{SWE-agent}: Agent-computer interfaces enable automated software engineering},
  author={Yang, John and Jimenez, Carlos E and Wettig, Alexander and Lieret, Kilian and Yao, Shunyu and Narasimhan, Karthik and Press, Ofir},
  journal={Advances in Neural Information Processing Systems},
  volume={37},
  pages={50528--50652},
  year={2024}
}

@inproceedings{jiang2023llmlingua,
  title={{LLMLingua}: Compressing prompts for accelerated inference of large language models},
  author={Jiang, Huiqiang and Wu, Qianhui and Lin, Chin-Yew and Yang, Yuqing and Qiu, Lili},
  booktitle={Proceedings of the 2023 conference on empirical methods in natural language processing},
  pages={13358--13376},
  year={2023}
}

@inproceedings{chevalier2023adapting,
  title={Adapting language models to compress contexts},
  author={Chevalier, Alexis and Wettig, Alexander and Ajith, Anirudh and Chen, Danqi},
  booktitle={Proceedings of the 2023 Conference on Empirical Methods in Natural Language Processing},
  pages={3829--3846},
  year={2023}
}

@article{packer2023memgpt,
  title={Memgpt: Towards llms as operating systems},
  author={Packer, Charles and Wooders, Sarah and Lin, Kevin and Fang, Vivian and Patil, Shishir G and Stoica, Ion and Gonzalez, Joseph E},
  journal={arXiv preprint arXiv:2310.08560},
  year={2023}
}

@inproceedings{park2023generative,
  title={Generative agents: Interactive simulacra of human behavior},
  author={Park, Joon Sung and O'Brien, Joseph and Cai, Carrie Jun and Morris, Meredith Ringel and Liang, Percy and Bernstein, Michael S},
  booktitle={Proceedings of the 36th annual acm symposium on user interface software and technology},
  pages={1--22},
  year={2023}
}

@inproceedings{besta2024graph,
  title={Graph of thoughts: Solving elaborate problems with large language models},
  author={Besta, Maciej and Blach, Nils and Kubicek, Ales and Gerstenberger, Robert and Podstawski, Michal and Gianinazzi, Lukas and Gajda, Joanna and Lehmann, Tomasz and Niewiadomski, Hubert and Nyczyk, Piotr and others},
  booktitle={Proceedings of the AAAI conference on artificial intelligence},
  volume={38},
  number={16},
  pages={17682--17690},
  year={2024}
}

@article{yao2023tree,
  title={Tree of thoughts: Deliberate problem solving with large language models},
  author={Yao, Shunyu and Yu, Dian and Zhao, Jeffrey and Shafran, Izhak and Griffiths, Tom and Cao, Yuan and Narasimhan, Karthik},
  journal={Advances in neural information processing systems},
  volume={36},
  pages={11809--11822},
  year={2023}
}

@inproceedings{liu2023repobench,
  title={Repobench: Benchmarking repository-level code auto-completion systems},
  author={Liu, Tianyang and Xu, Canwen and McAuley, Julian},
  booktitle={International Conference on Learning Representations},
  volume={2024},
  pages={47832--47850},
  year={2024}
}

@article{wei2022chain,
  title={Chain-of-thought prompting elicits reasoning in large language models},
  author={Wei, Jason and Wang, Xuezhi and Schuurmans, Dale and Bosma, Maarten and Xia, Fei and Chi, Ed and Le, Quoc V and Zhou, Denny and others},
  journal={Advances in neural information processing systems},
  volume={35},
  pages={24824--24837},
  year={2022}
}

@article{shinn2023reflexion,
  title={Reflexion: Language agents with verbal reinforcement learning},
  author={Shinn, Noah and Cassano, Federico and Gopinath, Ashwin and Narasimhan, Karthik and Yao, Shunyu},
  journal={Advances in neural information processing systems},
  volume={36},
  pages={8634--8652},
  year={2023}
}

@inproceedings{wang2025openhands,
  title={Openhands: An open platform for ai software developers as generalist agents},
  author={Wang, Xingyao and Li, Boxuan and Song, Yufan and Xu, Frank F and Tang, Xiangru and Zhuge, Mingchen and Pan, Jiayi and Song, Yueqi and Li, Bowen and Singh, Jaskirat and others},
  booktitle={International Conference on Learning Representations},
  volume={2025},
  pages={65882--65919},
  year={2025}
}

@article{xia2025agentless,
  title={Demystifying llm-based software engineering agents},
  author={Xia, Chunqiu Steven and Deng, Yinlin and Dunn, Soren and Zhang, Lingming},
  journal={Proceedings of the ACM on Software Engineering},
  volume={2},
  number={FSE},
  pages={801--824},
  year={2025},
  publisher={ACM New York, NY, USA}
}

@article{wang2025openhandssdk,
  title={The openhands software agent sdk: A composable and extensible foundation for production agents},
  author={Wang, Xingyao and Rosenberg, Simon and Michelini, Juan and Smith, Calvin and Tran, Hoang H and Nyst, Engel and Malhotra, Rohit and Zhou, Xuhui and Chen, Valerie and Brennan, Robert and others},
  journal={Proceedings of Machine Learning and Systems},
  volume={8},
  pages={1843--1855},
  year={2026}
}

@article{lindenbauer2025complexity,
  title={The Complexity Trap: Simple Observation Masking Is as Efficient as LLM Summarization for Agent Context Management},
  author={Lindenbauer, Tobias and Slinko, Igor and Felder, Ludwig and Bogomolov, Egor and Zharov, Yaroslav},
  journal={arXiv preprint arXiv:2508.21433},
  year={2025}
}

@article{kang2025acon,
  title={{ACON}: Optimizing context compression for long-horizon {LLM} agents},
  author={Kang, Minki and Chen, Wei-Ning and Han, Dongge and Inan, Huseyin A and Wutschitz, Lukas and Chen, Yanzhi and Sim, Robert and Rajmohan, Saravan},
  journal={arXiv preprint arXiv:2510.00615},
  year={2025}
}

@article{deng2025swebenchpro,
  title={Swe-bench pro: Can ai agents solve long-horizon software engineering tasks?},
  author={Deng, Xiang and Da, Jeff and Pan, Edwin and He, Yannis Yiming and Ide, Charles and Garg, Kanak and Lauffer, Niklas and Park, Andrew and Pasari, Nitin and Rane, Chetan and others},
  journal={arXiv preprint arXiv:2509.16941},
  year={2025}
}

@article{yang2025swesmith,
  title={Swe-smith: Scaling data for software engineering agents},
  author={Yang, John and Lieret, Kilian and Jimenez, Carlos and Wettig, Alexander and Khandpur, Kabir and Zhang, Yanzhe and Hui, Binyuan and Press, Ofir and Schmidt, Ludwig and Yang, Diyi},
  journal={Advances in Neural Information Processing Systems},
  volume={38},
  year={2026}
}

@article{tbench2025,
  title={Terminal-bench: A benchmark for ai agents in terminal environments},
  author={Team, The Terminal-Bench},
  journal={https://github. com/laude-institute/terminal-bench},
  year={2025}
}

@article{badertdinov2025swerebench,
  title={Swe-rebench: An automated pipeline for task collection and decontaminated evaluation of software engineering agents},
  author={Badertdinov, Ibragim and Golubev, Alexander and Nekrashevich, Maksim and Shevtsov, Anton and Karasik, Simon and Andriushchenko, Andrei and Trofimova, Maria and Litvintseva, Daria and Yangel, Boris},
  journal={Advances in Neural Information Processing Systems},
  volume={38},
  year={2026}
}

@article{le2025sweevo,
  title={SWE-EVO: Benchmarking coding agents in long-horizon software evolution scenarios},
  author={Le, Tue and Thai, Minh VT and Manh, Dung Nguyen and Nhat, Huy Phan and Bui, Nghi DQ},
  journal={arXiv preprint arXiv:2512.18470},
  year={2025}
}

@article{chhikara2025mem0,
  title={{Mem0}: Building Production-Ready {AI} Agents with Scalable Long-Term Memory},
  author={Chhikara, Prateek and Khant, Dev and Aryan, Saket and Singh, Taranjeet and Yadav, Deshraj},
  journal={arXiv preprint arXiv:2504.19413},
  year={2025}
}

\newpage
\appendix
\section{Additional Benchmark Details}
\label{appendix:additional_benchmarks}

This section details the benchmarks of Section~\ref{sec:setup} and gives per-run results.
\begin{prompt}[title={\thetcbcounter{} Benchmarks}]
\begin{itemize}[leftmargin=*,itemsep=1pt,topsep=2pt]
    \item \textbf{SWE-Bench Pro}~\citep{deng2025swebenchpro} (public release, 642 tasks in 11 repositories): all 237 Python tasks, plus three subsets.
    \begin{itemize}[leftmargin=*,itemsep=0pt,topsep=0pt]
        \item A \emph{large-change} subset of 30 Python tasks whose gold patch touches at least 4 files or 150 lines; none of these tasks appears in the 26-task sample.
        \item The benchmark's 51 hardest tasks (20 Python, 20 Go, 10 JavaScript, 1 TypeScript).
        \item The remaining 344 Go, JavaScript and TypeScript tasks, i.e.\ those in neither the hardest 51 nor a 30-task pilot sample.
    \end{itemize}
    \item \textbf{SWE-Bench Multilingual}~\citep{yang2025swesmith}: all 300 tasks in 9 languages. We first ran 50 of them, sampled at random (seed 42), and report that sample separately.
    \item \textbf{Terminal-Bench 2.0}~\citep{tbench2025}: all 89 tasks.
    \item \textbf{SWE-rebench}~\citep{badertdinov2025swerebench} (leaderboard set of 860 tasks): all 216 tasks whose gold patch touches at least 3 files or 100 lines.
    \item \textbf{SWE-EVO}~\citep{le2025sweevo}: all 48 tasks. Each task asks the agent to implement a release's changes from its release notes.
\end{itemize}
For reference, we also report a 26-task Python sample of SWE-Bench Pro.
\end{prompt}

\begin{prompt}[title={\thetcbcounter{} Setup}]
The agent model is Claude Sonnet 4.5, with a cost cap of \$3 per task for the agent model. Results with Kimi-K2.5 and GLM-5 are in Table~\ref{tab:backbone_baselines}. Both methods run in \textsc{SWE-agent} with the same system prompt, tools and instance template. For benchmarks other than SWE-Bench Verified, both methods use the same language-neutral template, which changes two lines of the instance prompt. The baseline keeps the $n{=}6$ most recent observations, as in the main experiments.

All experiments use the following lower-cost configuration:
\begin{itemize}[leftmargin=*,itemsep=1pt,topsep=2pt]
    \item Claude Haiku 4.5 writes the node summaries.
    \item Parents are scored by shared files, shared identifiers and recency instead of being chosen by an LLM call.
    \item The two most recent action--observation pairs are always kept in full.
    \item An observation that is not selected for the context is replaced by its own node summary, truncated to 300 characters.
\end{itemize}
This configuration was chosen on a small development subset of SWE-Bench Verified and the 26-task Pro sample. It was then kept fixed for every benchmark.

Both methods use the same test-based stopping check. When the agent's patch could be submitted, the check runs the repository tests that reference the changed functions, using the repository's own test command, both before and after the patch. It stops the run if the agent's reproduction passes and no such test newly fails. Terminal-Bench tasks have no repository test suite, so neither method uses the check there.

Grading uses each benchmark's own evaluation:
\begin{itemize}[leftmargin=*,itemsep=1pt,topsep=2pt]
    \item the SWE-Bench harness for Verified and Multilingual, and the harness released with SWE-rebench;
    \item each task's test script for SWE-Bench Pro;
    \item the task tests run in the agent's container after the agent finishes, for Terminal-Bench;
    \item the official evaluation script of SWE-EVO, which also reports a \emph{fix rate}: per task, the fraction of its fail-to-pass tests that the patch makes pass, or zero if any pass-to-pass test breaks, averaged over tasks.
\end{itemize}
\end{prompt}

\paragraph{Runs and reporting.}
Most configurations are run three times; the sliding window and \toolname{} are run five times on the 26-task Pro sample. We report resolve rates as mean $\pm$ standard deviation across runs. Paired results for SWE-rebench (216 tasks) and the remaining Pro Go/JS/TS tasks appear in Table~\ref{tab:additional_benchmarks_single}.

\paragraph{Significance tests.}
Across the 6 distinct task sets in Table~\ref{tab:main}, excluding the Pro large-change subset, \toolname{} significantly outperforms the sliding window (exact McNemar test on pooled outcomes, $p{<}10^{-6}$). A one-sided sign test treating each task set as one observation gives $p{=}0.016$. With the Kimi-K2.5 and GLM-5 evaluations included, \toolname{} leads on 9 of the 9 task sets.

\paragraph{Task subsets.}
Table~\ref{tab:subsets} breaks the results down by properties of the tasks that are fixed before looking at the outcomes: the size of the gold patch, the language and the repository. \toolname{} leads on both patch sizes in every benchmark and backbone, on six of the seven Multilingual languages and on six of the seven Pro repositories. Its gain is not consistently larger on large changes. Pooled over all 361 SWE-Bench Pro tasks, including Go, JavaScript and TypeScript, the gain is significant ($p{<}0.001$), as it is on the large-change tasks of all benchmarks ($p{<}10^{-4}$).

\begin{table*}[t]
\centering
\small
\setlength{\tabcolsep}{3.5pt}
\resizebox{\textwidth}{!}{%
\begin{tabular}{@{}l c r r c r r@{\hspace{16pt}}l c r r c r r@{}}
\toprule
\textbf{Subset} & \textbf{Tasks} & \textbf{SW} & \textbf{CW} & \textbf{Only} & \textbf{$p$} & \textbf{$p_{\text{Holm}}$} & \textbf{Subset} & \textbf{Tasks} & \textbf{SW} & \textbf{CW} & \textbf{Only} & \textbf{$p$} & \textbf{$p_{\text{Holm}}$} \\
\midrule
\multicolumn{7}{@{}l}{\textit{Pooled}} & \multicolumn{7}{@{}l}{\textit{Multilingual, by language}} \\
SWE-Bench Pro, all languages & 361 & 52.4 & \textbf{57.4} & 143 / 89 & ${<}0.001$ & ${<}0.001$ & C & 28--29 & 43.1 & \textbf{48.9} & 13 / 8 & 0.38 & 1.00 \\
Pro Python, three backbones & 558 & 60.8 & \textbf{68.3} & 262 / 136 & ${<}0.001$ & ${<}0.001$ & Go & 38--41 & 22.5 & \textbf{26.7} & 14 / 9 & 0.40 & 1.00 \\
Large change, all benchmarks & 566 & 38.0 & \textbf{42.3} & 188 / 114 & ${<}0.001$ & ${<}0.001$ & Java & 39--43 & 25.7 & \textbf{36.9} & 24 / 10 & 0.02 & 0.17 \\
Large change, three backbones & 815 & 42.9 & \textbf{48.8} & 314 / 169 & ${<}0.001$ & ${<}0.001$ & JavaScript / TypeScript & 29--34 & 42.6 & \textbf{43.0} & 9 / 8 & 1.00 & 1.00 \\
\multicolumn{7}{@{}l}{\textit{Gold-patch size}} & PHP & 42--43 & 57.8 & \textbf{61.0} & 11 / 7 & 0.48 & 1.00 \\
Pro Python, large & 123 & 64.2 & \textbf{69.6} & 43 / 23 & 0.02 & 0.17 & Ruby & 40--44 & 54.7 & \textbf{57.0} & 14 / 11 & 0.69 & 1.00 \\
Pro Python, small & 63 & 70.9 & \textbf{78.3} & 25 / 11 & 0.03 & 0.20 & Rust & 37--43 & \textbf{63.9} & 60.8 & 12 / 16 & 0.57 & 1.00 \\
Pro Python, Kimi-K2.5, large & 123 & 52.3 & \textbf{61.0} & 63 / 31 & 0.001 & 0.01 & \multicolumn{7}{@{}l}{\textit{SWE-Bench Pro, by repository}} \\
Pro Python, Kimi-K2.5, small & 63 & 57.7 & \textbf{66.1} & 36 / 20 & 0.04 & 0.26 & ansible & 63 & 70.4 & \textbf{74.6} & 22 / 14 & 0.24 & 0.55 \\
Pro Python, GLM-5, large & 123 & 57.2 & \textbf{67.8} & 63 / 24 & ${<}0.001$ & ${<}0.001$ & openlibrary & 56 & 54.2 & \textbf{61.3} & 22 / 10 & 0.05 & 0.20 \\
Pro Python, GLM-5, small & 63 & 70.4 & \textbf{73.0} & 32 / 27 & 0.60 & 0.94 & qutebrowser & 67 & 73.1 & \textbf{80.1} & 24 / 10 & 0.02 & 0.15 \\
Pro Go/JS/TS, large & 143 & 37.5 & \textbf{39.9} & 54 / 44 & 0.36 & 0.94 & NodeBB & 23 & 13.0 & \textbf{34.8} & 18 / 3 & 0.001 & 0.01 \\
Pro Go/JS/TS, small & 32 & 37.5 & \textbf{47.9} & 21 / 11 & 0.11 & 0.55 & flipt & 60 & \textbf{41.1} & 35.6 & 18 / 28 & 0.18 & 0.55 \\
Multilingual, large & 46--50 & 32.4 & \textbf{41.9} & 24 / 10 & 0.02 & 0.19 & vuls & 48 & 43.1 & \textbf{52.1} & 23 / 10 & 0.04 & 0.18 \\
Multilingual, small & 216--239 & 45.5 & \textbf{47.7} & 77 / 62 & 0.23 & 0.94 & navidrome & 43 & 38.8 & \textbf{39.5} & 15 / 14 & 1.00 & 1.00 \\
Verified, Kimi-K2.5, small & 95 & 67.7 & \textbf{70.9} & 31 / 22 & 0.27 & 0.94 & \multicolumn{7}{l}{} \\
\bottomrule
\end{tabular}%
}
\caption{Resolve rate (\%) on task subsets fixed before looking at the outcomes (Claude Sonnet 4.5 unless a backbone is named; mean over three runs, pooled rows weight every task run equally). Large change: the gold patch touches at least 3 files or 100 lines, the rule that defines the large-change sets; the large-change pools also include SWE-rebench and SWE-EVO. Subsets with fewer than 10 tasks are left out, among them the 3 large-change Verified tasks. \emph{Only}: tasks resolved only by CW / only by SW within the same run, pooled over runs; $p$: exact McNemar test; $p_{\text{Holm}}$: adjusted within each block. SW: sliding window; CW: \toolname{}.}
\label{tab:subsets}
\end{table*}

\paragraph{Task inclusion.}
Unless otherwise noted, comparisons include only tasks graded in every run of both methods. Environment setup failures leave 186 of 237 Pro Python tasks and exclude about half of the remaining Pro Go/JS/TS tasks. Many excluded images cannot support the SWE-agent runtime; for Go/JS/TS, these failures affect the same tasks for both methods. We also exclude tasks whose gold patches fail the benchmark tests on SWE-rebench and SWE-EVO. Empty patches count as unresolved. For Multilingual, we use the tasks graded for both methods within each run, yielding 288, 262, and 289 tasks out of 300.

\paragraph{API rate limits.}
Some task runs ended early because API rate limits persisted after retries. We re-ran these tasks with the same configuration, for every method.

\begin{table*}[t]
\centering
\scriptsize
\resizebox{\linewidth}{!}{%
\begin{tabular}{@{}lccccr@{}}
\toprule
\textbf{Benchmark} & \textbf{Tasks} & \textbf{Runs} & \textbf{Sliding Window} & \textbf{\toolname{}} & \textbf{$\Delta$} \\
\midrule
SWE-Bench Pro, Python (all) & 186 & 3 & 66.5 $\pm$ 2.2 & \textbf{72.6 $\pm$ 1.4} & +6.1 \\
SWE-Bench Pro, Python (26 sample) & 26 & 5 & 72.3 $\pm$ 4.2 & 72.3 $\pm$ 5.0 & 0.0 \\
SWE-Bench Pro, large-change & 22 & 3 & 60.6 $\pm$ 5.2 & \textbf{71.2 $\pm$ 6.9} & +10.6 \\
SWE-Bench Pro, hardest 51 & 28 & 3 & 19.0 $\pm$ 5.5 & 19.0 $\pm$ 5.5 & 0.0 \\
SWE-Bench Multilingual (all 300) & 262--289 & 3 & 43.2 $\pm$ 1.6 & \textbf{46.7 $\pm$ 1.6} & +3.5 \\
SWE-Bench Multilingual (50 sample) & 48 & 3 & 39.6 $\pm$ 5.5 & \textbf{46.5 $\pm$ 7.3} & +6.9 \\
SWE-EVO & 42 & 3 & 23.8 $\pm$ 2.4 & \textbf{31.7 $\pm$ 2.7} & +7.9 \\
Terminal-Bench 2.0 & 61 & 3 & 43.2 $\pm$ 4.1 & \textbf{44.8 $\pm$ 0.9} & +1.6 \\
\bottomrule
\end{tabular}%
}
\caption{Resolve rate (\%, mean $\pm$ std over runs) with Claude Sonnet 4.5 and a \$3 cap per task. \emph{Tasks}: tasks graded in every run of both methods. \emph{Runs}: sliding window / \toolname{} where they differ. $\Delta$: \toolname{} minus sliding window. Multilingual (all 300): the graded set differs between runs, so \emph{Tasks} gives the range over runs. Terminal-Bench runs use no stopping check on either side.}
\label{tab:additional_benchmarks}
\end{table*}

\begin{table*}[t]
\centering
\small
\begin{tabular}{@{}llcccccc@{}}
\toprule
\textbf{Benchmark} & \textbf{Method} & \textbf{R1} & \textbf{R2} & \textbf{R3} & \textbf{Mean $\pm$ Std} & \textbf{Steps} & \textbf{\$/task} \\
\midrule
\multirow{2}{*}{Pro, Python (all, 186)} & Sliding Window & 119 & 125 & 127 & 66.5 $\pm$ 2.2 & 56 & 2.44 \\
  & \toolname{} & 136 & 132 & 137 & \textbf{72.6 $\pm$ 1.4} & 46 & 2.77 \\
\midrule
\multirow{2}{*}{Pro, large-change (22)} & Sliding Window & 14 & 12 & 14 & 60.6 $\pm$ 5.2 & 61 & 2.62 \\
  & \toolname{} & 14 & 16 & 17 & \textbf{71.2 $\pm$ 6.9} & 50 & 2.98 \\
\midrule
\multirow{2}{*}{Pro, hardest 51 (28)} & Sliding Window & 5 & 7 & 4 & 19.0 $\pm$ 5.5 & 61 & 2.74 \\
  & \toolname{} & 4 & 5 & 7 & 19.0 $\pm$ 5.5 & 51 & 3.26 \\
\midrule
\multirow{2}{*}{Multilingual (48)} & Sliding Window & 21 & 20 & 16 & 39.6 $\pm$ 5.5 & 71 & 3.03 \\
  & \toolname{} & 22 & 26 & 19 & \textbf{46.5 $\pm$ 7.3} & 60 & 3.55 \\
\midrule
\multirow{2}{*}{Multilingual (all 300)} & Sliding Window & 122 & 118 & 122 & 43.2 $\pm$ 1.6 & 71 & 3.00 \\
  & \toolname{} & 131 & 127 & 133 & \textbf{46.7 $\pm$ 1.6} & 60 & 3.54 \\
\midrule
\multirow{2}{*}{SWE-EVO (42)} & Sliding Window & 9 & 10 & 11 & 23.8 $\pm$ 2.4 & 74 & 2.78 \\
  & \toolname{} & 14 & 12 & 14 & \textbf{31.7 $\pm$ 2.7} & 61 & 3.20 \\
\midrule
\multirow{2}{*}{Terminal-Bench 2.0 (61)} & Sliding Window & 26 & 29 & 24 & 43.2 $\pm$ 4.1 & 52 & 2.16 \\
  & \toolname{} & 27 & 27 & 28 & \textbf{44.8 $\pm$ 0.9} & 45 & 2.45 \\
\bottomrule
\end{tabular}
\caption{Resolved tasks per run (R1--R3), mean resolve rate, average agent steps per task, and average cost per task. The cost covers the agent model plus, for \toolname{}, the Haiku 4.5 summaries, so it can exceed the \$3 cap on the agent model. Multilingual (all 300): runs 1--3 are graded on 288, 262 and 289 tasks. Steps and cost are averaged over all task runs, including tasks outside the graded set.}
\label{tab:additional_benchmarks_runs}
\end{table*}

\begin{table*}[t]
\centering
\small
\begin{tabular}{@{}lccc@{}}
\toprule
\textbf{Run} & \textbf{Only \toolname{}} & \textbf{Only Sliding Window} & \textbf{McNemar $p$} \\
\midrule
R1 & 23 & 6 & 0.002 \\
R2 & 21 & 14 & 0.31 \\
R3 & 24 & 14 & 0.14 \\
All runs & 68 & 34 & 0.001 \\
\bottomrule
\end{tabular}
\caption{Paired comparison on the 186 SWE-Bench Pro Python tasks: tasks resolved by only one method within the same run, with an exact McNemar test. The last row pools all three runs; it treats repeated runs as independent pairs, so its $p$-value is approximate.}
\label{tab:additional_benchmarks_mcnemar}
\end{table*}

\begin{table}[t]
\centering
\small
\setlength{\tabcolsep}{3pt}
\resizebox{\columnwidth}{!}{%
\begin{tabular}{@{}l c c c r c c c@{}}
\toprule
\textbf{Benchmark} & \textbf{Tasks} & \textbf{SW} & \textbf{CW} & \textbf{$\Delta$} & \textbf{Only} & \textbf{$p$} & \textbf{Steps} \\
\midrule
SWE-rebench (216) & 212 & 27.2 & \textbf{30.3} & +3.1 & 49 / 29 & 0.031 & 65 / 54 \\
Pro Go/JS/TS (344) & 175 & 37.5 & \textbf{41.3} & +3.8 & 75 / 55 & 0.095 & 65 / 54 \\
\bottomrule
\end{tabular}%
}
\caption{Full sets, mean of three runs per method (Claude Sonnet 4.5, \$3 cap). \emph{Tasks}: tasks graded in every run of both methods, after removing tasks whose gold patch fails the benchmark's tests and, for Pro Go/JS/TS, tasks whose image cannot host the SWE-agent runtime. \emph{Only}: tasks resolved only by CW / only by SW within the same run, pooled over runs; $p$: exact McNemar test; \emph{Steps}: SW / CW. No run ended on an API rate limit. SW: sliding window; CW: \toolname{}.}
\label{tab:additional_benchmarks_single}
\end{table}

\paragraph{Resolve rates.}
Tables~\ref{tab:additional_benchmarks} and~\ref{tab:additional_benchmarks_single} report detailed results. On the 160 Pro Python tasks outside the development sample, \toolname{} retains its advantage over the sliding window (72.1\% vs.\ 65.6\%). Gains on the remaining benchmarks are generally smaller, and not all of them are individually significant.

\paragraph{Repeated reads and retained observations.}
On the 186 Pro Python tasks, \toolname{} reduces read commands that exactly repeat an earlier command from 2.3 to 1.2 per task with Claude Sonnet 4.5 and from 17.0 to 6.3 with Kimi-K2.5. Among the observations retained in full beyond the two most recent steps, 56--57\% are more than six steps old, with a median age of seven to eight steps across the two backbones.

\paragraph{Agent steps and cost.}
\toolname{} uses 13--18\% fewer agent steps on every benchmark (Table~\ref{tab:additional_benchmarks_runs}). However, many runs stop at the budget limit, including over 80\% on Multilingual and Pro Go/JS/TS, so average step counts also reflect per-step cost. Table~\ref{tab:steps_budget} therefore compares steps on task runs that both methods solve before exhausting the budget. Longer prompts offset the reduction in steps, leaving agent-model costs per task similar across methods. Haiku 4.5 summaries add \$0.35--0.52 per task.

\paragraph{Other baselines.}
Table~\ref{tab:baselines_full} provides the detailed results for Figure~\ref{fig:baselines}. On SWE-EVO, full-history retention (29.4\%, three runs) improves over the sliding window (23.8\%) but remains below \toolname{} (31.7\%). Table~\ref{tab:retrieval} extends the retrieval comparison to additional benchmarks.

\begin{table}[h]
\centering
\small
\setlength{\tabcolsep}{3pt}
\begin{tabular}{@{}l c r r r@{}}
\toprule
\textbf{Benchmark (backbone)} & \textbf{On budget} & \textbf{Pairs} & \textbf{SW} & \textbf{CW} \\
\midrule
Pro Python (Sonnet 4.5) & 62 / 56\% & 131 & 36 & 29 \\
Pro Python (Kimi-K2.5) & 53 / 44\% & 170 & 41 & 37 \\
Pro Python (GLM-5) & 58 / 52\% & 126 & 46 & 38 \\
SWE-EVO (Sonnet 4.5) & 78 / 77\% & 5 & 32 & 41 \\
Terminal-Bench (Sonnet 4.5) & 49 / 45\% & 55 & 35 & 33 \\
\bottomrule
\end{tabular}
\caption{Steps without the budget effect. A run that reaches the cost cap stops regardless of its progress, so its step count partly reflects the cost of each step. \emph{On budget}: share of task runs that stop at the cap (SW / CW). \emph{Pairs}: task runs that both methods solve in the same repetition before reaching the cap; \emph{SW} and \emph{CW} give their average steps on these pairs. SW: sliding window; CW: \toolname{}.}
\label{tab:steps_budget}
\end{table}

\begin{table}[h]
\centering
\small
\setlength{\tabcolsep}{3.5pt}
\begin{tabular}{@{}l c r@{\,}l r r@{}}
\toprule
\textbf{Method} & \textbf{Runs} & \multicolumn{2}{c}{\textbf{Resolve (\%)}} & \textbf{Steps} & \textbf{\$/task} \\
\midrule
\multicolumn{6}{@{}l}{\textit{Pro sample ($n$=26)}} \\
Sliding window & 5 & \textbf{72.3} & {\scriptsize$\pm$4.2} & 58 & 2.54 \\
Full history & 3 & 67.9 & {\scriptsize$\pm$5.9} & 34 & 2.63 \\
LLM summarization & 3 & 64.1 & {\scriptsize$\pm$8.0} & 69 & 3.51 \\
Retrieval (BM25) & 3 & \underline{71.8} & {\scriptsize$\pm$2.2} & 48 & 2.57 \\
Hybrid & 3 & 67.9 & {\scriptsize$\pm$11.8} & 53 & 2.66 \\
\toolname{} & 5 & \textbf{72.3} & {\scriptsize$\pm$5.0} & 46 & 2.82 \\
\midrule
\multicolumn{6}{@{}l}{\textit{Pro large-change ($n$=22)}} \\
Sliding window & 3 & 60.6 & {\scriptsize$\pm$5.2} & 61 & 2.62 \\
Full history & 3 & \underline{63.6} & {\scriptsize$\pm$0.0} & 34 & 2.67 \\
LLM summarization & 3 & 62.1 & {\scriptsize$\pm$5.2} & 65 & 3.22 \\
Retrieval (BM25) & 3 & 62.1 & {\scriptsize$\pm$6.9} & 47 & 2.48 \\
Hybrid & 3 & 59.1 & {\scriptsize$\pm$7.9} & 54 & 2.65 \\
\toolname{} & 3 & \textbf{71.2} & {\scriptsize$\pm$6.9} & 50 & 2.98 \\
\bottomrule
\end{tabular}
\caption{Context-management baselines with Claude Sonnet 4.5 and a \$3 cap per task (Figure~\ref{fig:baselines}). Resolve rate is the mean $\pm$ std over runs; for resolve rate, best in \textbf{bold}, second \underline{underlined}. \$/task is the total cost, including \toolname{}'s Haiku summaries, the summarizer calls of LLM summarization (estimated from the trajectories) and the checkpoint summaries of the hybrid \citep{lindenbauer2025complexity}.}
\label{tab:baselines_full}
\end{table}

\begin{table}[h]
\centering
\small
\setlength{\tabcolsep}{2.4pt}
\begin{tabular}{@{}l c c c c c c@{}}
\toprule
\textbf{Benchmark} & \textbf{Runs} & \textbf{SW} & \textbf{CW} & \textbf{Ret.} & \textbf{Only} & \textbf{$p$} \\
\midrule
Pro Python (186) & 3 & 66.5 & \textbf{72.6} & \underline{69.4} & 56 / 38 & 0.079 \\
Multilingual (48) & 3 & \underline{39.6} & \textbf{46.5} & 38.9 & 22 / 11 & 0.08 \\
SWE-EVO (42) & 3 & 23.8 & \underline{31.7} & \textbf{32.5} & 14 / 15 & 1 \\
Terminal-Bench (58) & 3 & 42.0 & \underline{43.1} & \textbf{44.3} & 7 / 9 & 0.8 \\
\bottomrule
\end{tabular}
\caption{Resolve rate (\%) of the retrieval baseline (Ret.) with Claude Sonnet 4.5 and a \$3 cap, mean over runs, on the task sets of Table~\ref{tab:additional_benchmarks_runs}; best in \textbf{bold}, second \underline{underlined}. \emph{Only}: tasks resolved by only \toolname{} (CW) / only retrieval within the same run, pooled over runs; $p$: exact McNemar test. SW: sliding window.}
\label{tab:retrieval}
\end{table}

\section{Configuration Details}
\begin{prompt}[title={\thetcbcounter{} Parameters and Configuration},label=app:parameters]
In the experiments, parents are scored by shared files, shared identifiers and recency; the full framework (Table~\ref{tab:full_variant}) instead uses the LLM prompt in Appendix~\ref{prompt:memory_weaver_parent}. The context keeps $W{=}5$ earlier steps (ten history entries) in full besides the current step, two of them always the two most recent steps, so six observations in total; a candidate's parent score is $3\times$(shared files) $+\sqrt{\text{shared identifiers}}$ $+\,0.5\times$recency (its rank among the candidates, scaled to $[0,1]$), candidates need a shared file or at least three shared identifiers, and the five highest-scoring ones become parents (the root node if none qualifies). The full framework changes only how parents are chosen; traversal, the six full observations and the summary stand-ins are the same; parents are chosen among the 50 most recent valid nodes. Each node may attach to multiple parents, resulting in a directed acyclic graph. Dependency summaries are used only for dependency analysis and are never injected directly into the final agent context. Validation labels are derived from test execution outcomes. Nodes corresponding to failed executions are excluded from future parent selection but are retained as supporting evidence unless explicitly superseded. No instance-specific tuning is performed. In the selection controls (Table~\ref{tab:selection}), the three earlier observations are the most recent steps or random earlier steps (seeded per step).
\end{prompt}

\begin{prompt}[title={\thetcbcounter{} LLM Summarization and Hybrid Baselines},label=appendix:llm_summarization]

Our LLM summarization baseline periodically condenses older history into an LLM-generated summary while retaining the 6 most recent observation pairs intact. The same backbone model performs both the task and the summarization. All other settings are identical to the sliding-window baseline. The hybrid follows \citet{lindenbauer2025complexity} with their published hyperparameters ($N{=}43$, $M{=}10$, masking window 10) and their summary prompt.
\end{prompt}

\section{Observation Compression for Non-selected Entries}
\label{appendix:observation-compress-sec}

For history entries outside the selected dependency subgraph, \toolname{} compresses only the \textbf{Observation} ($O_i$) component while preserving the corresponding \textbf{Thought} ($T_i$) and \textbf{Action} ($A_i$). Compression replaces the original observation with a lightweight placeholder that records the amount of omitted content, computed directly from the original observation; this mirrors the truncation of the sliding-window baseline. In the low-cost configuration used in our experiments, the placeholder is replaced by the step's own node summary, truncated to 300 characters.

\begin{prompt}[title={\thetcbcounter{} Compressed Observations},label=appendix:observation-compress,unbreakable]
\promptsubsection{Placeholder}
\begin{lstlisting}[style=markdown]
Old environment output: (N lines omitted) (M images omitted)
\end{lstlisting}
\promptsubsection{Low-cost configuration}
\begin{lstlisting}[style=markdown]
Earlier result (summary): <node summary>
\end{lstlisting}
\end{prompt}

\section{Case Studies}
\label{appendix:earlier_eval}

\label{sec:case_study}
We looked for tasks that one method solves in every run and the other in none: on the 186 Pro Python tasks, two with Claude Sonnet 4.5, four with Kimi-K2.5 and five with GLM-5 favor \toolname{}, and one, two and two favor the sliding window. Appendix~\ref{appendix:case_studies} gives step-level excerpts.

\begin{prompt}[title={\thetcbcounter{} The sliding window revisits what it dropped},unbreakable]
In \texttt{qutebrowser-9b71c1ea} (Kimi-K2.5; \toolname{} 3/3 runs, sliding window 0/3), the task adds a locale workaround that touches \texttt{qtargs.py} and \texttt{configdata.yml}. The sliding-window agent reads the relevant parts of both files within its first 30 steps. Because only the last six observations stay in its context, in one run it then repeats the same searches for the locale setting in \texttt{configdata.yml} again and again: 33 of its 110 steps repeat an earlier read command verbatim, and it reaches the cost cap without editing a file. \toolname{} keeps the views that the current step depends on. It reads each range once, runs the tests at step 25, edits both files from step 30, verifies and submits at step 60.
\end{prompt}

\begin{prompt}[title={\thetcbcounter{} The sliding window circles the tests},unbreakable]
In \texttt{openlibrary-2fe532a3} (GLM-5; \toolname{} 3/3, sliding window 0/3), in one run the sliding-window agent keeps re-running the same four searches of \texttt{test\_vendors.py}: 32 of its 88 steps repeat an earlier read command verbatim, and it reaches the cost cap without editing a file. \toolname{} reads the relevant parts of \texttt{vendors.py} between steps 31 and 50 and keeps those views in full until it edits the file at step 76, up to 45 steps later; its patch passes the tests.
\end{prompt}

\begin{prompt}[title={\thetcbcounter{} When recency wins},unbreakable]
The sliding window wins two of three runs of \texttt{qutebrowser-ebfe9b7} and \toolname{} none, a refactor that moves Qt's message handler from \texttt{log.py} to \texttt{qtlog.py}. Both methods edit the same two files. In one \toolname{} run, the change to \texttt{log.py} replaces an import that the rest of the file still uses. Here the latest full view of the file being edited matters more than older dependencies, which \toolname{} keeps only as summaries.
\end{prompt}

\section{Detailed Case Study Traces}
\label{appendix:case_studies}
Excerpts from the runs discussed in Appendix~\ref{sec:case_study}. ``(=$k$)'' marks a command identical to the one at step $k$.

\begin{prompt}[title={\thetcbcounter{} Case 1: qutebrowser-9b71c1ea, Kimi-K2.5},label=prompt:case_study]
\promptsubsection{Sliding window (unresolved, 110 steps, cost cap)}
\texttt{2}~~view qtargs.py \quad \texttt{4}~~view test\_qtargs.py \quad \texttt{11--14}~~view qtargs.py in ranges \quad \texttt{16}~~view configdata.yml 295--320\\
\texttt{17}~~\texttt{grep -rn "workarounds.locale"} \quad \texttt{32}~~\texttt{grep -rn "locale" config/}\\
\texttt{74}~~view configdata.yml 295--320 (=16) \quad \texttt{76}~~\texttt{grep -rn "locale" config/} (=32) \quad \texttt{78}~~\texttt{grep -rn "workarounds.locale"} (=17)\\
\texttt{81--109}~~the same searches in a cycle: \texttt{grep "workarounds"} and \texttt{grep "locale"} on configdata.yml, \texttt{grep -rn "locale"} on the package, \texttt{grep -rn "workarounds.locale"}, view configdata.yml 295--320\\
No source file is edited.

\promptsubsection{ContextWeaver (resolved, 62 steps)}
\texttt{2}~~view qtargs.py \quad \texttt{5}~~view test\_qtargs.py \quad \texttt{17}~~view configdata.yml 295--330 \quad \texttt{19--22}~~view qtargs.py in ranges\\
\texttt{25}~~run \texttt{pytest tests/unit/config/test\_qtargs.py} \quad \texttt{27--28}~~write and run a reproduction script\\
\texttt{30--31}~~edit qtargs.py \quad \texttt{33}~~edit configdata.yml \quad \texttt{36--39}~~refine qtargs.py and re-run the script\\
\texttt{40}~~run the tests \quad \texttt{57--60}~~inspect the diff and submit\\
One read command in the run repeats an earlier one.
\end{prompt}

\section{Implementation Details}
\label{sec:appendix}

\paragraph{Integration with runtime validation}
After constructing the context, \toolname{} prepends a summary of the latest test results to the agent's next input. This allows the agent to see which tests passed or failed without retrieving the earlier test output.

\begin{prompt}[title={\thetcbcounter{} Prepended Test Summary},label=prompt:test_summary,unbreakable]
\small
\texttt{TEST STATUS:}\\
\texttt{~~\checkmark~test\_basic: PASSED}\\
\texttt{~~\ding{55}~test\_login\_validation: FAILED}\\
\texttt{---}\\
\texttt{Q: Fix the authentication bug}
\end{prompt}
\begin{prompt}[title={\thetcbcounter{} Prompt for \toolname Parent Selection (full framework)},label=prompt:memory_weaver_parent]
\promptsubsection{System}
\begin{lstlisting}[style=markdown]
You maintain the dependency graph of a software-engineering agent's trajectory. Each step is a node, and you link the latest step to the earlier steps it directly builds on: the file view it edits from or reasons about, the test or error output it responds to, the search result or definition it relies on. The agent's context later follows these links back through earlier dependencies, so pick the direct dependencies precisely; the steps they in turn build on are reached through their own links. The files in focus are listed for you, and each candidate says which file it read, if any. Prefer the most recent step that shows a given file region. Return a JSON list of 1 to 3 step ids, most important first, e.g. [12, 7]. Return [] only if the step builds on no earlier step.
\end{lstlisting}
\noindent\textbf{User:}
\begin{lstlisting}[style=markdown]
Task (truncated):
{problem statement}

Earlier steps (id, action -> start of output):
[0] (full read of lib/parse.py) 'str_replace_editor view /app/lib/parse.py' -> '...'
...
[24] (kept in full) 'python reproduce.py' -> 'Traceback ...'

Latest step 25:
thought: ...   action: ...   output: ...
Files in focus (latest thought and last five actions, newest first): [...]

Which earlier steps (ids from the list) does the latest step directly build on? Answer with the JSON list only.
\end{lstlisting}
\end{prompt}

\section{Tool Use and Prompt Setup for Agent}
\begin{prompt}[title={\thetcbcounter{} Prompt and Tools for SWE-agent Setup},label=prompt:agent_setup]
\promptsubsection{System Template}
\begin{lstlisting}[style=markdown]
You are a helpful assistant that can interact with a computer to solve tasks.
\end{lstlisting}

\promptsubsection{Instance Template}
\begin{lstlisting}[style=markdown]
<uploaded_files>
{{working_dir}}
</uploaded_files>
I've uploaded a python code repository in the directory {{working_dir}}.
Consider the following PR description:

<pr_description>
{{problem_statement}}
</pr_description>

Can you help me implement the necessary changes to the repository so that
the requirements specified in the <pr_description> are met?
I've already taken care of all changes to any of the test files described
in the <pr_description>. This means you DON'T have to modify the testing
logic or any of the tests in any way!
Your task is to make the minimal changes to non-tests files in the
{{working_dir}} directory to ensure the <pr_description> is satisfied.
Follow these steps to resolve the issue:
1. As a first step, it might be a good idea to find and read code relevant
   to the <pr_description>
2. Create a script to reproduce the error and execute it with
   `python <filename.py>` using the bash tool, to confirm the error
3. Edit the sourcecode of the repo to resolve the issue
4. Rerun your reproduce script and confirm that the error is fixed!
5. Think about edgecases and make sure your fix handles them as well
Your thinking should be thorough and so it's fine if it's very long.
\end{lstlisting}

\promptsubsection{Next Step Template}
\begin{lstlisting}[style=markdown]
OBSERVATION:
{{observation}}
\end{lstlisting}

\promptsubsection{Tools Configuration}
\begin{lstlisting}[style=markdown]
execution_timeout: 300
bundles:
  - tools/registry
  - tools/edit_anthropic
  - tools/review_on_submit_m
  - tools/diff_state
enable_bash_tool: true
parse_function: function_calling
\end{lstlisting}

\promptsubsection{Submit Review Message}
\begin{lstlisting}[style=markdown]
Thank you for your work on this issue. Please carefully follow the steps
below to help review your changes.

1. If you made any changes to your code after running the reproduction
   script, please run the reproduction script again. If the reproduction
   script is failing, please revisit your changes and make sure they are
   correct. If you have already removed your reproduction script, please
   ignore this step.
2. Remove your reproduction script (if you haven't done so already).
3. If you have modified any TEST files, please revert them to the state
   they had before you started fixing the issue. You can do this with
   `git checkout -- /path/to/test/file.py`. Use below <diff> to find the
   files you need to revert.
4. Run the submit command again to confirm.

Here is a list of all of your changes:

<diff>
{{diff}}
</diff>
\end{lstlisting}
\end{prompt}

\section{Algorithm Walkthrough}
\label{appendix:walkthrough}
\begin{algorithm}[H]
\small
\caption{\toolname: Dependency-Aware Context Construction}
\label{alg:contextweaver}
\begin{algorithmic}[1]
\REQUIRE History $\mathbf{H}=\{H_0,\ldots,H_k\}$, query $Q$, graph $G=(V,E)$
\REQUIRE Hyper-parameters: window $W$ (steps), max parents $m$%, candidate limit $L$
\ENSURE Context $C$, updated graph $G$
\STATE $N_k \leftarrow \phi(H_k, Q)$
  \hfill $\triangleright$ \textit{Node extraction}

\STATE $\mathcal{C}_k \leftarrow$ nodes not marked
  \textsc{Failed} or \textsc{Superseded}

\FOR{each candidate $N_i \in \mathcal{C}_k$}
  \STATE $p_i \leftarrow$ estimated likelihood that $N_k$ depends on $N_i$
\ENDFOR

\STATE $S_k \leftarrow$ top-$m$ nodes from $\mathcal{C}_k$ ranked by $p_i$
  \hfill $\triangleright$ \textit{Parents}

\STATE $G \leftarrow \big(V \cup \{N_k\},\;
  E \cup \{N_i \!\to\! N_k : N_i \in S_k\}\big)$

\IF{$|\mathbf{H}| \leq W$}
  \STATE \textbf{return} $\mathbf{H}$, $G$
    \hfill $\triangleright$ \textit{Warmup: use full history}
\ENDIF

% \vspace{0.3em}
\STATE $A \leftarrow \{N_k\}$;\;\; $\mathcal{Q} \leftarrow S_k$
  \hfill $\triangleright$ \textit{Ancestry (BFS)}
\WHILE{$\mathcal{Q} \neq \varnothing \wedge |A| \leq W$}
  \STATE Pop node $N$ from $\mathcal{Q}$
  \STATE $A \leftarrow A \cup \{N\}$
  \STATE Enqueue all parents of $N$ in $G$ not already in $A$
\ENDWHILE

% \vspace{0.3em}
\STATE $C \leftarrow [\;]$
  \hfill $\triangleright$ \textit{Weave the context}
\FOR{$i = 0, \ldots, k$}
  \IF{$N_i \in A$}
    \STATE Append $H_i$ to $C$
      \hfill $\triangleright$ \textit{Keep ancestor in full}
  \ELSE
    \STATE Append $\mathrm{Compress}(H_i)$ to $C$
      \hfill $\triangleright$ \textit{Compress}
  \ENDIF
\ENDFOR

\STATE \textbf{return} $C$, $G$
\end{algorithmic}
\end{algorithm}
\begin{prompt}[title={\thetcbcounter{} Algorithm Walkthrough:\allowbreak\ pytest-dev\_\_pytest-\allowbreak5262},label=prompt:algorithm_walkthrough]

\promptsubsection{Bug}

\begin{lstlisting}[style=markdown]
- EncodedFile reports mode as "rb+" (binary)
- write() only accepts strings (not bytes)
- Callers infer binary mode and send bytes -> TypeError
\end{lstlisting}

\promptsubsection{History Indexing Convention}

\begin{lstlisting}[style=markdown]
- Each tool interaction produces two history entries:
  (1) Thought + Action
  (2) Observation
- Therefore the history index increases by 2 each time:
  H_2, H_4, H_6, ...
- Nodes are only constructed from action-aligned history entries
\end{lstlisting}

\promptsubsection{Node Under Analysis}

\begin{lstlisting}[style=markdown]
- We analyze node N_18, constructed from history entry H_18
- At this point, the agent has applied a fix and is about to validate it
\end{lstlisting}

\promptsubsection{Line 1: Node Extraction}

\begin{lstlisting}[style=markdown]
Input (H_18):
- Thought: "__getattr__ delegates mode to buffer returning 'rb+'"
- Action: edit /testbed/src/_pytest/capture.py
- Observation: added @property stripping 'b'

Output (N_18):
- id: trace_5262_18
- summary: override delegated mode with explicit property
- parents: unset
- dependency_summary: unset
- validation: unknown
\end{lstlisting}

\promptsubsection{Line 2: Parent Selection}

\begin{lstlisting}[style=markdown]
Candidate nodes:
- N_6: view capture.py
- N_8: inspect EncodedFile
- N_12: analyze __getattr__ delegation
- N_14: build reproduction script
- N_16: confirm failure ('rb+' -> TypeError)

Selected parents:
- N_12 (root cause analysis)
- N_16 (failure confirmation)

Parents(N_18) = {N_12, N_16}
\end{lstlisting}

\promptsubsection{Line 3: Ancestry Extraction}

\begin{lstlisting}[style=markdown]
Backward chains:
- N_12 -> N_8 -> N_6
- N_16 -> N_14 -> N_8 -> N_6

Merged ancestry set:
- P_18 = {N_6, N_8, N_12, N_14, N_16}

Dependency summaries incrementally encode:
- locating module
- inspecting EncodedFile
- tracing __getattr__ delegation
- reproducing failure
- applying fix
\end{lstlisting}

% TODO: if validation filtering is added back, restore this section
% \promptsubsection{Line 4: Validation Filtering}
%
% \begin{lstlisting}[style=markdown]
% Validation states:
% - unknown: N_6, N_8, N_12, N_14
% - failed:  N_16
%
% Rule:
% - failed nodes are kept as evidence
% - only superseded nodes are removed
%
% Filtered set:
% - {N_6, N_8, N_12, N_14, N_16}
% \end{lstlisting}

\promptsubsection{Line 4: Context Selection}

\begin{lstlisting}[style=markdown]
- |H| = 18, warmup window W = 5 => Phase 2 (Weave)

Full T-A-O detail retained for:
- N_6, N_8, N_12, N_14, N_16, N_18

History alignment:
- N_6  -> H_6
- N_8  -> H_8
- N_12 -> H_12
- N_14 -> H_14
- N_16 -> H_16
- N_18 -> H_18

Compressed (observation elided):
- H_1-H_5, H_7, H_9-H_11, H_13, H_15, H_17
\end{lstlisting}
\end{prompt}

\end{document}